%% file: main.tex
\documentclass[]{rtp_tech_report}

\usepackage{algorithm}
\usepackage{algpseudocode}
\usepackage{wrapfig}
\usepackage{needspace}
\usepackage{afterpage}

\graphicspath{{figures/}}

\title{SCOPE: Subspace Clustering with Online Per-Head Top-K Estimation for Sparse Video Attention}

\author[1,\ast,\ddagger]{Qi Zhao}
\author[1,\ast,\ddagger]{Qirui Li}
\author[2,\dagger]{Hanlin Tang}
\author[2]{Yiduo Li}
\author[2]{Zhen Guo}
\author[2]{Cuifeng Shen}
\author[2]{Chao Xu}
\author[2]{Zhaosheng Chi}
\author[2]{Xiaojin Lu}
\author[2]{Kan Liu}
\author[2]{Tao Lan}
\author[2]{Lin Qu}
\author[1,\mathsection]{Xi Li}

\affiliation[1]{Zhejiang University}
\affiliation[2]{Alibaba Group}

\paperemail{qizhao@zju.edu.cn, qirui.l@zju.edu.cn, wanghaisheng.whs@alibaba-inc.com}
\firstpagenotes{%
  \textsuperscript{*} Equal contribution
  \hfill
  \textsuperscript{\textdagger} Project lead
  \hfill
  \textsuperscript{\S} Corresponding author
  \hfill
  \textsuperscript{\textdaggerdbl} Work done during internship at Alibaba Group
}

\abstract{
\input{sections/abstract}
}

\begin{document}

\maketitle

\begin{figure}[H]
    \vspace{-8mm}
    \centering
    \captionsetup{font=footnotesize,skip=0pt}
    \includegraphics[width=0.83\textwidth]{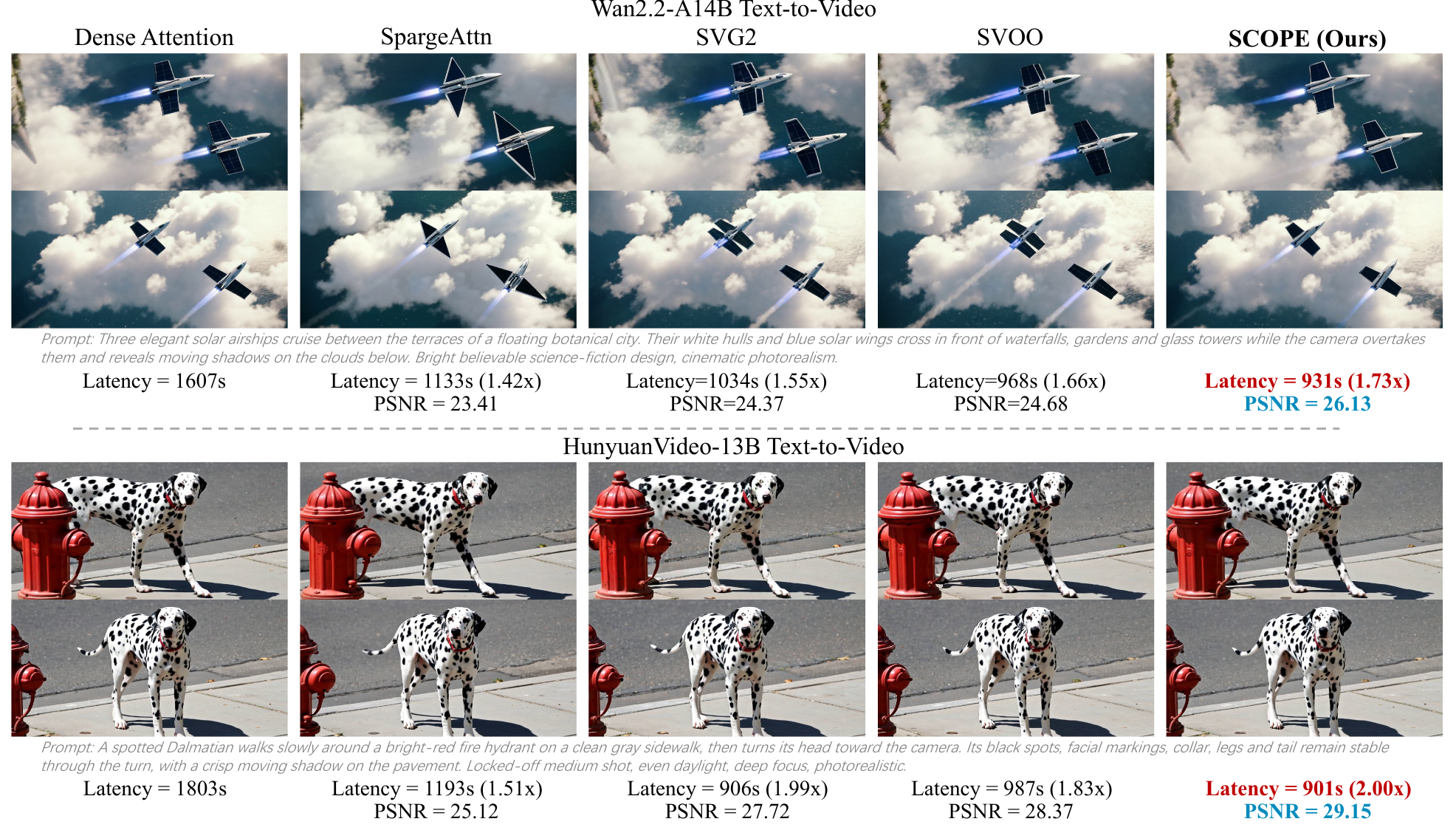}
    \caption{
    Representative 720p T2V runs on Wan2.2-A14B and HunyuanVideo-13B. SCOPE achieves significant speedup while maintaining fidelity to dense attention. The displayed metrics are per-run.
    }
    \label{fig:first-figure}
\end{figure}

\input{sections/introduction}
\afterpage{\input{sections/early_floats}}
\input{sections/related_work}
\input{sections/method}
\input{sections/experiment}
\input{sections/conclusion}

\bibliographystyle{plainnat}
\bibliography{references}

\clearpage
\beginappendix

\input{supplementary/sections/scope_appendix_sections_A_B}

\end{document}

%% file: sections/abstract.tex
Diffusion Transformers (DiTs) incur quadratic self-attention
cost over spatiotemporal tokens. Existing training-free
sparse attention methods often construct sparse masks from
block-level or cluster-level proxy scores, which can obscure
fine-grained differences among keys and miss high contribution
keys under aggressive sparsity. Moreover, such proxy scores
may yield overly concentrated softmax distributions, causing
Top-$p$ to retain too few keys for some query clusters.
Although a fixed Top-$k$ minimum alleviates this failure mode,
a shared value cannot adapt to variations across heads and
inputs. To address both limitations, we propose SCOPE, a
training-free sparse attention framework that combines
3D-RoPE-aligned key subspace clustering with online per-head
Top-$k$ estimation for efficient video-DiT inference.
SCOPE partitions post-RoPE keys into temporal, height, and
width subspaces, clusters them independently, and aggregates
the corresponding centroid scores through lookup tables to
obtain per key proxy scores for each query cluster.
Building on existing hybrid Top-$p$/fixed Top-$k$ selection,
SCOPE derives a head-specific Top-$k$ value online by
averaging the initial retained key counts within each head,
weighted by query cluster size, and selects additional keys
only for query clusters whose initial retained key counts fall
below this value. Sparse attention is then computed over the
selected original keys and values. Across six model--task
configurations, SCOPE consistently outperforms existing
training-free baselines in both fidelity and latency, achieving
up to a $1.99\times$ end-to-end speedup on 720p HunyuanVideo
with $28.46$ dB PSNR relative to dense attention.

%% file: sections/introduction.tex
\section{Introduction}
\label{sec:Introduction}

Modern video Diffusion Transformers (DiTs) flatten
spatiotemporal latents into tens of thousands of
tokens~\cite{dit,ma2024latte,yang2025cogvideox,
kong2024hunyuanvideo,wan2025wan}, making the
$O(N^2d)$ cost of self-attention a major bottleneck at high
resolutions and long video durations. Training-free sparse
attention reduces this cost by evaluating only selected
query--key interactions without modifying the pretrained
model. An effective sparse selector must therefore answer
two coupled questions: which keys matter to each query
group and how many to retain so that sparse attention remains faithful to the
dense-attention reference.

Existing training-free methods avoid constructing the dense
score matrix through predefined sparsity
patterns~\cite{STA,radial,LVSA,sparsevdit,CalibAtt},
block-level proxies~\cite{xu2025xattention,
zhang2025spargeattention,shen2025draftattention,adaspa,
hu2026dfsattn}, or clustering-based
proxies~\cite{svg2,svoo,tan2026adacluster,
lee2026hypervattention}. The latter two reduce online selection cost by representing
multiple keys with a pooled block or cluster centroid.
However, keys sharing the same representative receive the
same estimated importance, obscuring key-level differences and potentially omitting
influential keys whose importance is not reflected by the
shared proxy.

A natural way to recover this lost granularity is to exploit
the 3D rotary position embeddings (RoPE) used by video
DiTs~\cite{su2024roformer,wei2025videorope}. Temporal,
height and width coordinates act on disjoint channel ranges,
whose post-RoPE key subspaces exhibit distinct clustering
patterns, as shown in Figure~\ref{fig:rope-subspace-cluster}.
Representing each key with one full-dimensional assignment
couples these heterogeneous groupings, modeling the three
ranges separately preserves their distinct structures,
yielding finer-grained key discrimination.

Reliable masks also require calibrated retained key counts. Top-$p$ adapts to the estimated distribution but may under-select when approximation over-concentrates it. A fixed Top-$k$ floor helps~\cite{zhang2026spargeattention2}, but one global value cannot capture variations across heads and inputs. Offline head-wise schedules require dense calibration~\cite{svoo} and remain static, motivating an online, input-adaptive estimation.

Based on these observations, we propose \textbf{SCOPE}
(\textbf{S}ubspace \textbf{C}lustering with
\textbf{O}nline \textbf{P}er-Head Top-$k$
\textbf{E}stimation), a training-free sparse attention
framework for accelerating video-DiT inference. SCOPE keeps
query clustering full-dimensional, while partitioning each post-RoPE key into temporal, height and width ranges of 3D RoPE and clustering them independently, following the compositional principle of product quantization~\cite{pq}. Query-centroid slices score the corresponding subspace centroids to form compact lookup tables, each key retrieves and sums one entry per table to obtain its proxy score. These compositional scores improve key discrimination without expensive query--key scoring and are used only for selection, sparse attention is computed with the original tokens.

For count estimation, SCOPE applies hybrid Top-$p$/fixed-Top-$k$ selection to each query cluster, sets the online per-head Top-$k$ to the query-size-weighted average of the initial selected counts and extends only those query clusters below this value. Larger selections remain unchanged, so no offline dense profiling or stored head-wise schedules are required.

\begin{figure}[t]
    \centering
    \includegraphics[width=0.82\columnwidth]{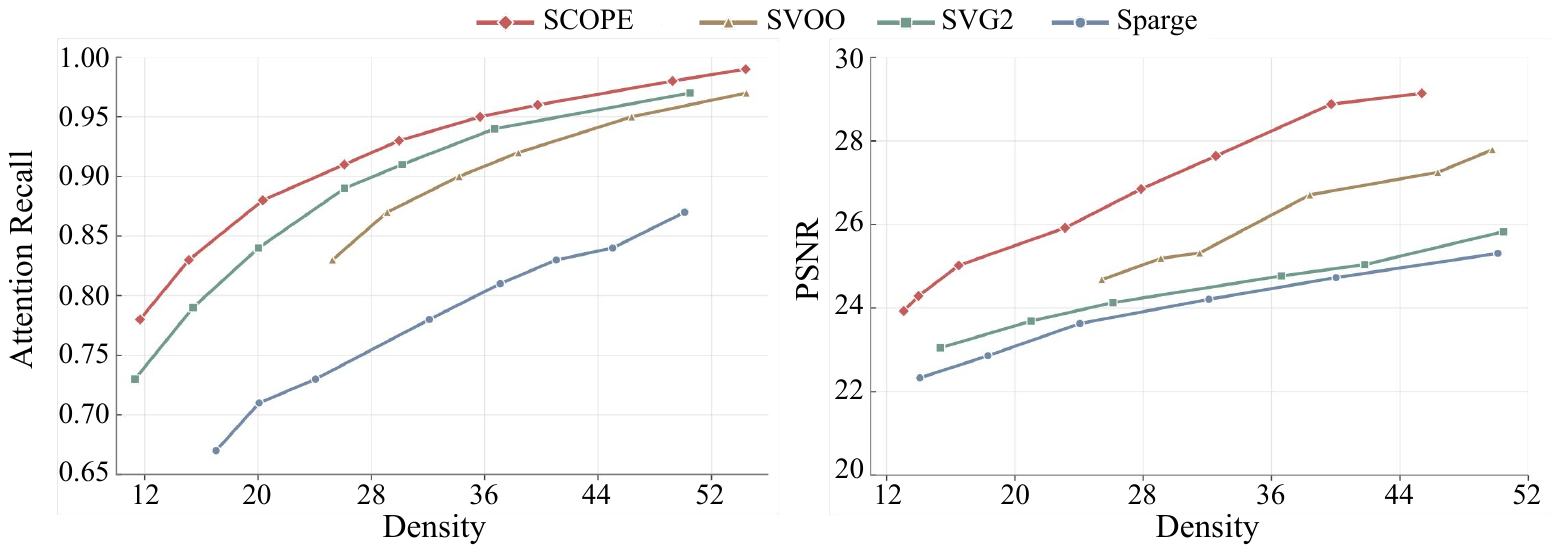}
    \caption{
Attention recall and PSNR under matched attention density.
SCOPE consistently achieves higher attention recall and PSNR
than competing sparse attention methods at the same retained
density.
}
\label{fig:recall}
\end{figure}

Experiments across six 720p model--task settings spanning
text-to-video and image-to-video generation on Wan2.1,
Wan2.2 and HunyuanVideo demonstrate consistent
improvements in fidelity and latency over training-free
sparse attention baselines. In particular, SCOPE achieves up
to a $1.99\times$ end-to-end speedup on
HunyuanVideo-T2V, with $28.46$ dB PSNR relative to the
corresponding dense-attention output. Representative T2V comparisons in
Figure~\ref{fig:first-figure} further illustrate this
fidelity--efficiency trade-off. At matched attention density,
Figure~\ref{fig:recall} further shows that SCOPE
consistently achieves the highest attention recall and PSNR
across the evaluated densities, with especially clear gains
in PSNR. These results indicate that SCOPE uses the same
sparse attention budget more effectively. Our main contributions
are:

\begin{itemize}
    \item We propose 3D-RoPE-aligned key subspace
    clustering, which independently clusters the temporal,
    height and width channel ranges and composes their
    centroid scores into fine-grained key-level proxy logits
    with additive scoring cost.

    \item We introduce online per-head Top-$k$ estimation,
    which derives an adaptive retention floor from base
    counts weighted by query cluster size and extends only
    clusters below this floor, without offline dense
    calibration.

    \item We evaluate SCOPE on six 720p T2V and I2V
    settings across three video DiT families. SCOPE achieves
    the best dense-reference fidelity and lowest measured
    latency among the evaluated training-free baselines,
    delivering $1.67\times$--$1.99\times$ end-to-end
    speedups.
\end{itemize}

%% file: sections/early_floats.tex
\begin{figure*}[!t]
    \centering
    \includegraphics[width=0.85\textwidth]{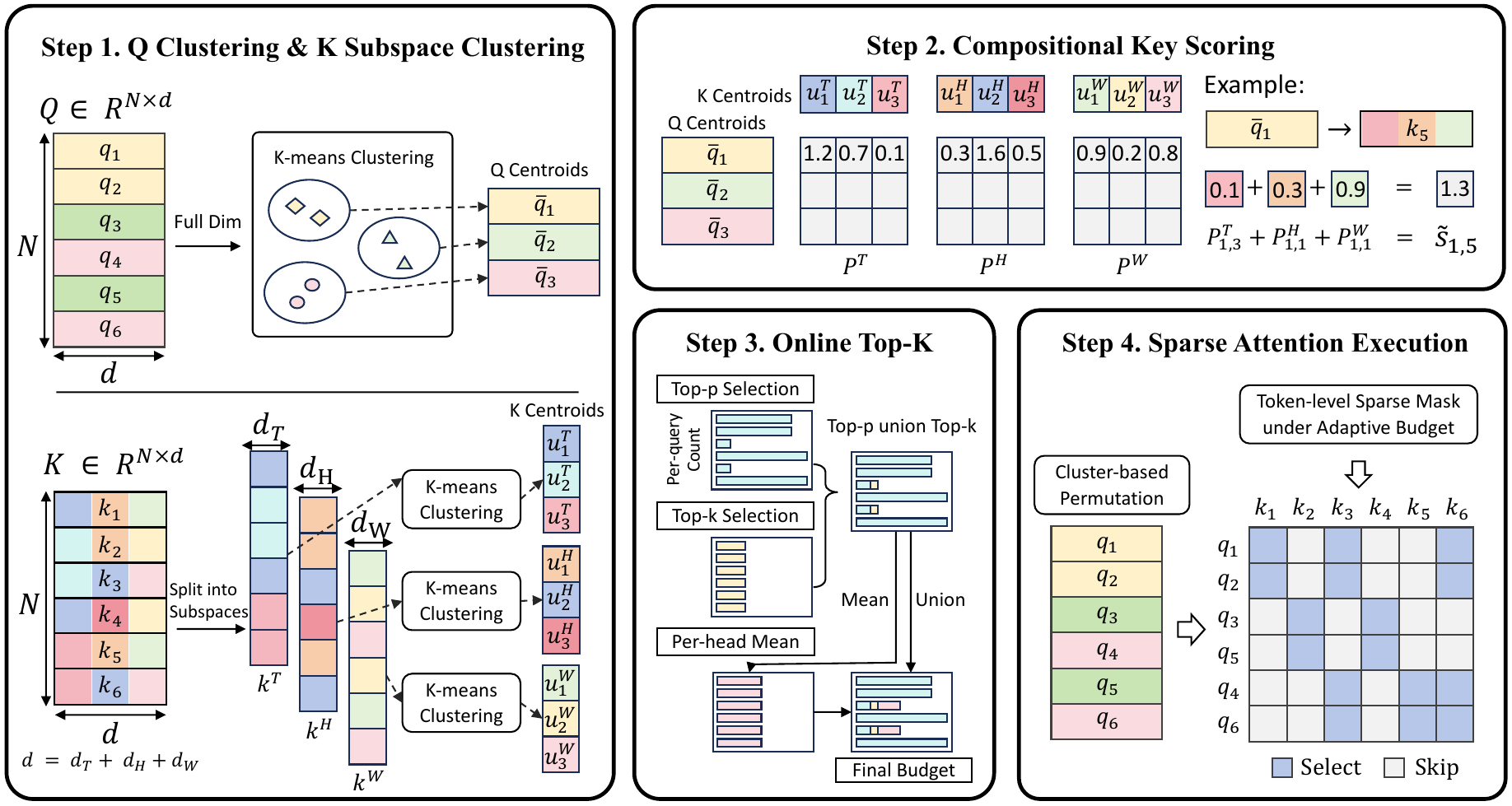}
    \caption{
Overview of SCOPE.
Queries are clustered in the full feature space, while each
post-RoPE key is represented by temporal, height and width
subspace assignments. Query-centroid slices score the
corresponding subspace centroids to form lookup tables, and
indexed summation reconstructs a proxy logit for every key.
Hybrid Top-$p$/fixed-Top-$k$ selection produces the initial
key counts, from which SCOPE estimates an online per-head
Top-$k$ value. Sparse attention is computed using the selected
original $K$ and $V$.
}
    \label{fig:method}
\end{figure*}

%% file: sections/related_work.tex
\section{Related Work}

\subsection{Predefined and Offline-Calibrated Sparse Masks}
Predefined methods restrict attention using local windows,
sliding tiles, or distance-aware layouts, while
offline-calibrated methods exploit cross-input regularities
to derive sparse
configurations~\cite{STA,LVSA,hassani2026generalized,
radial,sparsevdit,compaceattn,CalibAtt,
zhou2026scalingattention}. These approaches incur little
online mask-construction overhead and often admit regular,
hardware-friendly execution. However, masks determined
before observing the current activations may miss
prompt-dependent long-range interactions, nonlocal motion,
or abrupt scene changes.

\subsection{Dynamic Sparse Attention for Video Generation}

Dynamic methods infer sparse attention patterns from current
activations, adapting selection to the input and denoising
state~\cite{wu2025vmoba,
sun2026vorta,adaspa,hu2026dfsattn,rainfusion,
shmilovich2025liteattention,liu2026mixture,
long2026dynamicrad}. Some use lightweight approximations of
current attention~\cite{svg,vsa,xu2025xattention,shen2025draftattention,zhang2025spargeattention}, while others group or reorder tokens by
semantic similarity or query--key
affinity~\cite{svg2,tan2026adacluster,svoo,
zhao2026paroattention,ren2025grouping,RainFusion2},
complementary work improves sparse execution, reduces
approximation error, or adapts retained-key
counts~\cite{qiao2025flashomni,liu2025astraea,svgear,
liu2025rectifiedspa,chen2026re,feng2025quantsparse,
li2026pisa,pasa,zheng2026haste}.
SpargeAttention2~\cite{zhang2026spargeattention2} studies
hybrid Top-$p$/Top-$k$ selection in a trainable setting.
Yet many practical methods represent multiple keys with one
pooled block or full-dimensional centroid,
forcing heterogeneous keys to share a proxy score. SCOPE
instead combines 3D-RoPE-aligned subspace codewords into
per-key proxy scores and estimates a head-specific Top-$k$
online from current query-cluster statistics, without
retraining or offline dense calibration.

\subsection{Subspace Decomposition for Efficient Attention}

Product quantization~\cite{pq} decomposes vectors into
low-dimensional subspaces with independent codebooks. In
LLM inference, it supports KV-cache
compression~\cite{zhang2025pqcache} and codebook-based
query--key score approximation through indexed
accumulation~\cite{zhang2024nomad,
karmore2026lookat,yang2026self,
song2026csattention}. FASA~\cite{wang2026fasa} and
RTPurbo~\cite{zhou2026full} also estimate key relevance in
reduced feature spaces, using informative RoPE frequency
chunks and a learned low-dimensional indexer with
query-dependent Top-$p$ selection, respectively. These
approaches mainly target autoregressive LLMs. For diffusion
transformers, RoPeSLR~\cite{liu2026ropeslr} combines
3D-RoPE-aware sparse attention with low-rank approximation.
SCOPE instead keeps query clustering full-dimensional,
independently clusters post-RoPE keys along the temporal,
height and width channel ranges of 3D RoPE, and sums the
resulting subspace scores for fine-grained per-key ranking in
bidirectional video DiTs. Sparse attention then uses the
selected original keys and values.

%% file: sections/method.tex
\section{Method}
\label{sec:method}

Figure~\ref{fig:method} presents an overview of SCOPE.
Given post-RoPE queries and keys, SCOPE first clusters the
queries in the full feature space, allowing all queries in a
cluster to share one proxy ranking of the keys. It then splits
each key along the temporal, height and width channel ranges
of 3D RoPE and clusters the three subspaces independently.
The corresponding slices of each query centroid score the
subspace codebooks and indexed summation reconstructs a
proxy logit for every key. Finally, SCOPE combines hybrid
Top-$p$/fixed-Top-$k$ selection with an online per-head
Top-$k$ estimate and computes sparse attention using the
selected original keys and values.

\subsection{Problem Formulation}

Consider one self-attention head at a fixed layer and denoising
step. Let $Q,K\in\mathbb{R}^{N\times d}$ denote the query and
key matrices after 3D RoPE, with rows $q_i$ and $k_j$, and let
$V\in\mathbb{R}^{N\times d}$ denote the value matrix. Dense
attention computes
\begin{equation}
    O
    =
    \operatorname{softmax}\!\left(
    \frac{QK^\top}{\sqrt d}
    \right)V,
    \label{eq:dense-attention}
\end{equation}
which evaluates all $N^2$ query--key interactions. Our goal is
to construct a retained key set for each query cluster without
materializing the dense score matrix. SCOPE amortizes key
selection through full-dimensional query clustering and
approximates query-centroid--key logits with 3D-RoPE-aligned
key codebooks. The original queries and selected original keys
and values are then used for sparse attention.

\subsection{Full-Dimensional Query Clustering}
\label{sec:query-grouping}

Following clustering-based sparse attention
methods~\cite{svg2,svoo,tan2026adacluster}, we partition the
post-RoPE queries into $C_q\ll N$ clusters using K-means. Let
$g_i\in\{1,\ldots,C_q\}$ denote the cluster assignment of
$q_i$. For cluster $c$, define
\begin{equation}
\begin{aligned}
    \mathcal{Q}_c
    &=
    \{i:g_i=c\},
    &
    n_c
    &=
    |\mathcal{Q}_c|,
    &
    \bar q_c
    &=
    \frac{1}{n_c}
    \sum_{i\in\mathcal{Q}_c}q_i.
\end{aligned}
\label{eq:query-cluster}
\end{equation}
All queries in $\mathcal{Q}_c$ share the same proxy ranking
and selected key set $\mathcal{S}_c$, reducing the number of
independently constructed rankings from $N$ to $C_q$. The
centroid $\bar q_c$ is used only for key selection, the final
sparse attention uses the original query vectors.

Queries are not clustered independently in the RoPE
subspaces. Instead, each full-dimensional centroid $\bar q_c$
is sliced only when scoring the corresponding key codebooks.
This preserves one query grouping while allowing the key
subspaces to form different cluster assignments.

\begin{figure}[!t]
    \centering
    \includegraphics[
        width=0.82\columnwidth
    ]{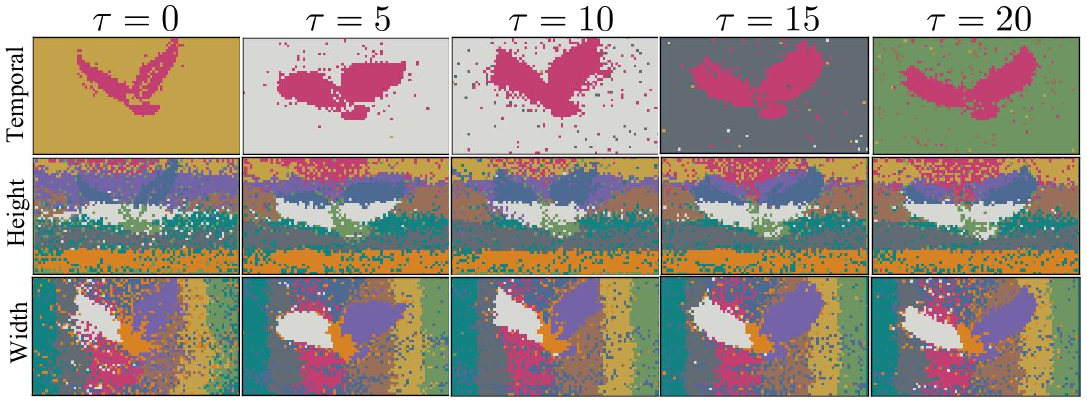}
    \caption{
Cluster assignments in 3D-RoPE key subspaces.
Post-RoPE keys on a $21\times45\times80$ grid are
clustered independently over the temporal, height and width
channel ranges with $C_{\mathrm{vis}}=10$, columns show
temporal positions $\tau\in\{0,5,10,15,20\}$. Height and
width assignments exhibit predominantly horizontal and
vertical structures, while temporal assignments vary more
across frames. Variations near foreground regions further
indicate content-dependent structure. Results are shown for
Wan2.2-T2V-A14B at layer 30, head 10 and denoising
step 35. Colors are comparable across columns only within
each row.
}
    \label{fig:rope-subspace-cluster}
\end{figure}

\subsection{3D-RoPE-Aligned Key Subspace Scoring}
\label{sec:subspace-scoring}

In 3D RoPE, the rotary transformations for temporal, height,
and width coordinates act on three disjoint channel ranges.
SCOPE follows these native ranges to split each post-RoPE
key $k_j$ into $k_j^{\mathrm{T}}$, $k_j^{\mathrm{H}}$, and
$k_j^{\mathrm{W}}$. Each full-dimensional query centroid
$\bar q_c$ is sliced in the same way only when scoring the
corresponding key subspaces. As illustrated in
Figure~\ref{fig:rope-subspace-cluster}, the three post-RoPE
key subspaces exhibit distinct clustering patterns, motivating
them to be modeled independently.

For each subspace
$m\in\{\mathrm{T},\mathrm{H},\mathrm{W}\}$, SCOPE applies
K-means to $\{k_j^m\}_{j=1}^{N}$ and obtains $C_m$
subspace centroids, denoted by
$u_1^m,\ldots,u_{C_m}^m$. The assignment of key $j$ in
subspace $m$ is
\begin{equation}
    z_j^m
    =
    \underset{1\le r\le C_m}{\arg\min}\,
    \left\|k_j^m-u_r^m\right\|_2^2.
    \label{eq:key-subspace-clustering}
\end{equation}
The three assignments jointly form the composite code of
key $j$:
\begin{equation}
    \mathbf{z}_j
    =
    \left(
    z_j^{\mathrm{T}},
    z_j^{\mathrm{H}},
    z_j^{\mathrm{W}}
    \right).
    \label{eq:key-composite-code}
\end{equation}
Unlike a single full-dimensional cluster assignment,
$\mathbf{z}_j$ allows the three RoPE subspaces to organize
keys independently. Keys assigned to the same centroid in
one subspace can still be distinguished by their assignments
in the other subspaces.

To score the keys for a query cluster, SCOPE compares each
query-centroid slice only with the centroids in the
corresponding key subspace. For each subspace $m$, these
partial dot products form a compact score table
$P_m\in\mathbb{R}^{C_q\times C_m}$, whose entry is
\begin{equation}
    P_m[c,r]
    =
    \left\langle
    \bar q_c^m,u_r^m
    \right\rangle.
    \label{eq:subspace-score-table}
\end{equation}
The columns of $P_m$ correspond to subspace centroids rather
than individual key tokens. The assignment $z_j^m$ maps
key $j$ to one column of each table. Retrieving and summing
the three corresponding entries gives its proxy logit:
\begin{equation}
    \tilde s_{c,j}
    =
    \frac{1}{\sqrt d}
    \sum_{m\in\{\mathrm{T},\mathrm{H},\mathrm{W}\}}
    P_m[c,z_j^m].
    \label{eq:lookup-scoring}
\end{equation}
Applying Eq.~\eqref{eq:lookup-scoring} to all query clusters
and keys produces
$\tilde S=[\tilde s_{c,j}]\in\mathbb{R}^{C_q\times N}$,
the query-cluster--key proxy-score matrix used for subsequent
key selection.

This compositional scoring provides a favorable trade-off
between computation and key discrimination. For each query
cluster, constructing the three compact tables requires only
$C_{\mathrm{T}}+C_{\mathrm{H}}+C_{\mathrm{W}}$ centroid dot
products, while the independent codebooks admit up to
$C_{\mathrm{T}}C_{\mathrm{H}}C_{\mathrm{W}}$ composite key
representations. Once the tables are constructed, each key
score requires only three indexed reads and two additions.
Thus, the number of expensive dot products grows additively
with the codebook sizes, whereas the number of possible
composite representations grows multiplicatively.

Because dot products are additive across disjoint channel
ranges, summing the three partial scores introduces no
additional approximation. The key-side approximation comes
from replacing each key slice with its assigned subspace
centroid. The resulting proxy scores are used only for key
selection, while the subsequent sparse attention is computed
using the selected original keys and values.

\subsection{Online Per-Head Top-$k$ Estimation}
\label{sec:online-topk}

Given the proxy-logit matrix $\tilde S$ obtained in
Eq.~\eqref{eq:lookup-scoring}, SCOPE first determines a
base retained key count for each query cluster and then
derives an online minimum for the current attention head.

\paragraph{Hybrid Top-$p$/fixed-Top-$k$ selection.}
Let $\tilde{\boldsymbol{s}}_c$ denote the $c$-th row of
$\tilde S$, and let
$\tilde{\boldsymbol{a}}_c=
\operatorname{softmax}(\tilde{\boldsymbol{s}}_c)$ be the
corresponding normalized proxy distribution. We use
$\tilde a_{c,j}$ to denote the probability assigned to key
$j$. Let $\pi_c$ be a permutation of the key indices that
orders them by decreasing proxy logit. Given a Top-$p$
threshold $\rho\in(0,1]$, the retained key count is
\begin{equation}
    t_c^{(p)}
    =
    \min\left\{
    t:
    \sum_{\ell=1}^{t}
    \tilde a_{c,\pi_c(\ell)}
    \ge \rho
    \right\}.
    \label{eq:top-p}
\end{equation}
Approximation errors may make the normalized proxy
distribution overly concentrated, causing Top-$p$ to retain
too few keys. We therefore combine it with a fixed Top-$k$
ratio $\alpha\in(0,1]$:
\begin{equation}
    k_{\mathrm{fix}}=\lceil\alpha N\rceil,
    \qquad
    b_c=\max\left\{t_c^{(p)},k_{\mathrm{fix}}\right\}.
    \label{eq:base-key-count}
\end{equation}
\input{sections/table_t2v}

Because both selections are prefixes of the same ranking
$\pi_c$, their union contains exactly $b_c$ keys.

\paragraph{Online per-head estimation.}
The fixed minimum prevents severe under-selection but cannot
account for variations across attention heads and inputs.
SCOPE instead derives a head-specific minimum from the base
counts observed in the current head:
\begin{equation}
    k_{\mathrm{head}}
    =
    \left\lceil
    \frac{
    \sum_{c=1}^{C_q} n_c b_c
    }{
    \sum_{c=1}^{C_q} n_c
    }
    \right\rceil.
    \label{eq:online-topk}
\end{equation}
Before rounding, Eq.~\eqref{eq:online-topk} is the average
base retained key count over all queries in the current head.
Weighting by $n_c$ makes each query, rather than each query
cluster, contribute equally to the estimate. The final
retained key count and selected key set are
\begin{equation}
    r_c=\max\left\{b_c,k_{\mathrm{head}}\right\},
    \qquad
    \mathcal{S}_c=
    \{\pi_c(1),\ldots,\pi_c(r_c)\}.
    \label{eq:final-key-set}
\end{equation}
Clusters with $b_c\ge k_{\mathrm{head}}$ remain unchanged,
while those below the head-specific minimum are extended
along the same proxy-logit ranking. The estimate is derived
entirely from the current head and input, requiring no offline
dense-attention profiling or stored head-wise schedules.

\input{sections/table_i2v}

\paragraph{Sparse attention execution.}
After $\mathcal{S}_c$ has been constructed, all queries in
$\mathcal{Q}_c$ reuse the same selected key set. Although
the key set is shared, each query computes its own attention
weights using its original representation. Let
$Q_{\mathcal{Q}_c}$ denote the corresponding query rows,
and let $K_{\mathcal{S}_c}$ and $V_{\mathcal{S}_c}$ denote
the selected rows of the original key and value matrices.
SCOPE computes
\begin{equation}
    O_{\mathcal{Q}_c}
    =
    \operatorname{softmax}\!\left(
    \frac{
    Q_{\mathcal{Q}_c}K_{\mathcal{S}_c}^{\top}
    }{\sqrt d}
    \right)
    V_{\mathcal{S}_c}.
    \label{eq:scope-execution}
\end{equation}
Once $\mathcal{S}_c$ is determined, the query centroids,
subspace centroids, assignments and proxy logits are no
longer used in the attention computation. The approximation
is therefore confined to selecting the retained interactions,
their attention logits and value aggregation are computed
from the original tokens. SCOPE requires
neither retraining nor modification of the pretrained model
parameters.

%% file: sections/table_t2v.tex
\begin{table*}[t]
    \centering
    \small
    \setlength{\tabcolsep}{3pt}
    \renewcommand{\arraystretch}{1.08}
    \resizebox{\textwidth}{!}{%
    \begin{tabular}{@{}llccccccccc@{}}
        \toprule
        Model & Baseline & PSNR $\uparrow$ & SSIM $\uparrow$ & LPIPS $\downarrow$ & ImageQual $\uparrow$ & AesQual $\uparrow$ & SubConsist $\uparrow$ & BackConsist $\uparrow$ & Latency & Speedup \\
        \midrule
        \multirow{5}{*}{\shortstack[l]{Wan 2.1-14B-\\T2V-720P}}
        & Full   & -- & -- & -- & 70.03\% & 59.06\% & 96.01\% & 96.31\% & 1913s & 1.00x \\
        & SpargeAttn & 23.86 & 0.791 & 0.1461 & 69.78\% & \textbf{59.09\%} & \textbf{95.94\%} & \textbf{96.21\%} & 1293s & 1.48x \\
        & SVG2   & 24.63 & 0.804 & 0.1268 & \textbf{70.12\%} & 58.76\% & 95.58\% & 95.84\% & 1211s & 1.58x \\
        & SVOO   & 25.71 & 0.831 & 0.1083 & 69.93\% & 59.06\% & 95.22\% & 95.57\% & 1117s & 1.71x \\
        & \textbf{Ours} & \textbf{26.11} & \textbf{0.844} & \textbf{0.1004} & 69.96\% & 59.07\% & 95.93\% & 96.14\% & \textbf{1085s} & \textbf{1.76x} \\
        \midrule
        \multirow{5}{*}{\shortstack[l]{Wan 2.2-A14B-\\T2V-720P}}
        & Full   & -- & -- & -- & 71.18\% & 61.56\% & 95.16\% & 95.77\% & 1599s & 1.00x \\
        & SpargeAttn & 24.37 & 0.813 & 0.1231 & 70.98\% & 61.48\% & \textbf{95.12\%} & 95.75\% & 1138s & 1.41x \\
        & SVG2   & 24.85 & 0.829 & 0.1128 & 71.06\% & 61.53\% & 94.92\% & 95.64\% & 1036s & 1.54x \\
        & SVOO   & 25.42 & 0.837 & 0.1018 & 71.02\% & 61.48\% & 94.98\% & 95.62\% & 966s & 1.66x \\
        & \textbf{Ours} & \textbf{26.01} & \textbf{0.859} & \textbf{0.0953} & \textbf{71.08\%} & \textbf{61.57\%} & 95.09\% & \textbf{95.76\%} & \textbf{930s} & \textbf{1.72x} \\
        \midrule
        \multirow{5}{*}{HunyuanVideo-T2V}
        & Full   & -- & -- & -- & 66.46\% & 57.80\% & 95.74\% & 96.20\% & 1801s & 1.00x \\
        & SpargeAttn & 24.54 & 0.809 & 0.1631 & 66.50\% & \textbf{58.06\%} & 95.61\% & 96.04\% & 1197s & 1.50x \\
        & SVG2   & 27.41 & 0.845 & 0.1067 & 65.06\% & 56.96\% & 95.48\% & 95.88\% & 907s & 1.99x \\
        & SVOO   & 28.08 & 0.871 & 0.0971 & 65.99\% & 57.58\% & 95.66\% & 96.09\% & 991s & 1.82x \\
        & \textbf{Ours} & \textbf{28.46} & \textbf{0.878} & \textbf{0.0927} & \textbf{66.55\%} & 57.82\% & \textbf{95.68\%} & \textbf{96.11\%} & \textbf{904s} & \textbf{1.99x} \\
        \bottomrule
    \end{tabular}%
    }
    \caption{
720p text-to-video results. PSNR, SSIM and LPIPS are measured
against dense attention, latency and speedup are end to end. Best
sparse method results are bold.
}
    \label{tab:main-results}
\end{table*}

%% file: sections/table_i2v.tex
\begin{table*}[t]
    \centering
    \small
    \setlength{\tabcolsep}{3pt}
    \renewcommand{\arraystretch}{1.08}
    \resizebox{\textwidth}{!}{%
    \begin{tabular}{@{}llccccccccc@{}}
        \toprule
        Model & Baseline & PSNR $\uparrow$ & SSIM $\uparrow$ & LPIPS $\downarrow$ & ImageQual $\uparrow$ & AesQual $\uparrow$ & SubConsist $\uparrow$ & BackConsist $\uparrow$ & Latency & Speedup \\
        \midrule
        \multirow{5}{*}{\shortstack[l]{Wan 2.1-14B-\\I2V-720P}}
        & Full   & -- & -- & -- & 72.09\% & 60.95\% & 95.16\% & 95.63\% & 1672s & 1.00x \\
        & SpargeAttn & 23.14 & 0.722 & 0.1484 & 72.11\% & 60.68\% & \textbf{95.02\%} & 95.59\% & 1121s & 1.49x \\
        & SVG2   & 23.87 & 0.757 & 0.1341 & \textbf{72.27\%} & 60.37\% & 94.32\% & 95.11\% & 1063s & 1.57x \\
        & SVOO   & 24.82 & 0.787 & 0.1153 & 72.11\% & 60.72\% & 94.22\% & 95.24\% & 997s & 1.68x \\
        & \textbf{Ours} & \textbf{26.77} & \textbf{0.838} & \textbf{0.0854} & 72.15\% & \textbf{60.78\%} & 94.96\% & \textbf{95.61\%} & \textbf{966s} & \textbf{1.73x} \\
        \midrule
        \multirow{5}{*}{\shortstack[l]{Wan 2.2-A14B-\\I2V-720P}}
        & Full   & -- & -- & -- & 72.20\% & 62.21\% & 96.56\% & 96.42\% & 1628s & 1.00x \\
        & SpargeAttn & 24.79 & 0.788 & 0.1022 & 72.12\% & 62.08\% & 96.42\% & 96.31\% & 1143s & 1.42x \\
        & SVG2   & 25.31 & 0.805 & 0.0984 & \textbf{72.29\%} & 61.62\% & 96.17\% & 96.10\% & 1085s & 1.50x \\
        & SVOO   & 27.38 & 0.849 & 0.0706 & 72.21\% & 62.15\% & 96.32\% & \textbf{96.37\%} & 1021s & 1.59x \\
        & \textbf{Ours} & \textbf{27.76} & \textbf{0.865} & \textbf{0.0665} & 72.16\% & \textbf{62.18\%} & \textbf{96.43\%} & 96.32\% & \textbf{974s} & \textbf{1.67x} \\
        \midrule
        \multirow{5}{*}{HunyuanVideo-I2V}
        & Full   & -- & -- & -- & 72.53\% & 60.76\% & 96.99\% & 96.51\% & 1783s & 1.00x \\
        & SpargeAttn & 22.68 & 0.778 & 0.1243 & \textbf{72.56\%} & \textbf{60.68\%} & 97.03\% & \textbf{96.56\%} & 1154s & 1.55x \\
        & SVG2   & 23.06 & 0.796 & 0.1183 & 72.11\% & 60.14\% & 96.91\% & 96.25\% & 1029s & 1.73x \\
        & SVOO   & 23.34 & 0.813 & 0.1168 & 72.06\% & 60.50\% & 96.94\% & 96.43\% & 1091s & 1.63x \\
        & \textbf{Ours} & \textbf{24.11} & \textbf{0.837} & \textbf{0.0962} & \textbf{72.56\%} & 60.57\% & \textbf{97.04\%} & 96.45\% & \textbf{1022s} & \textbf{1.74x} \\
        \bottomrule
    \end{tabular}%
    }
    \caption{
720p image-to-video results. PSNR, SSIM and LPIPS are measured
against dense attention, latency and speedup are end to end. Best
sparse method results are bold.
    }
    \label{tab:i2v-results}

    \vspace{8pt}
    \includegraphics[width=\textwidth]{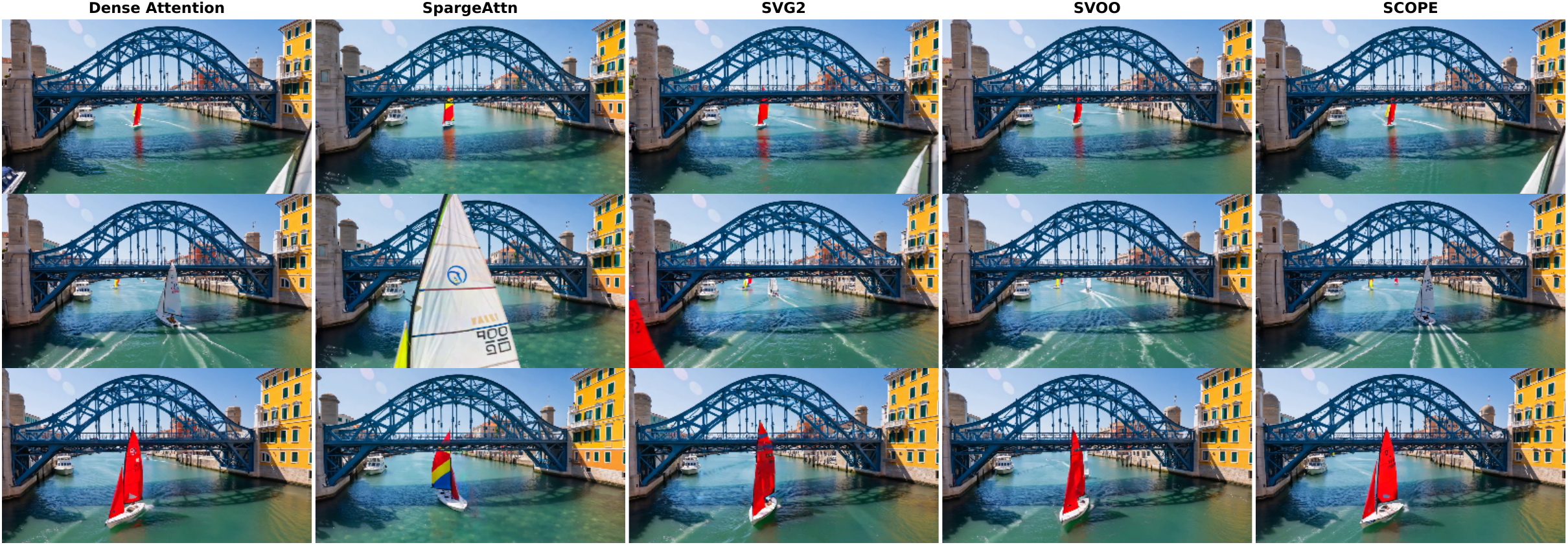}
    \captionsetup{font=small,skip=3pt}
    \captionof{figure}{
Qualitative comparison.
Representative frames generated using identical prompt and seed. SCOPE preserves the appearance and structure
of the dense-attention reference more faithfully than the
competing sparse attention methods.
}
    \label{fig:quality-compare}
\end{table*}

%% file: sections/experiment.tex
\Needspace{8\baselineskip}
\section{Experiment}

\subsection{Experimental Setup}

\paragraph{Models and datasets.}
We evaluate T2V and I2V generation on Wan2.1-14B, Wan2.2-A14B and HunyuanVideo-13B, yielding six model--task settings. T2V uses Penguin Benchmark prompts released with HunyuanVideo, and I2V uses VBench++~\cite{huang2024vbench++}. All videos are generated at $1280\times720$, I2V inputs are cropped to $16{:}9$.

\paragraph{Baselines and metrics.}
We compare SCOPE with three training-free sparse attention
methods: SpargeAttn~\cite{zhang2025spargeattention},
SVG2~\cite{svg2} and SVOO~\cite{svoo}, using dense
attention as the reference. We adopt their official
implementations and default configurations, adjusting only
exposed sparsity controls for the reported operating points. 
All methods share model weights, conditioning inputs, random seeds and sampling configurations. PSNR, SSIM and LPIPS measure fidelity to dense outputs. ImageQual, AesQual, SubConsist and BackConsist from VBench~\cite{huang2024vbench} assess generation quality. Efficiency is reported as end-to-end latency and speedup over dense attention.

\paragraph{Implementation details.}
All experiments are conducted on NVIDIA H200 GPUs. All K-means operations use the same implementation as SVG2~\cite{svg2}.
SCOPE uses $C_q=300$ query clusters and
$C_{\mathrm{T}}=C_{\mathrm{H}}=C_{\mathrm{W}}=333$
centroids for the temporal, height and width key subspaces.
The global fixed Top-$k$ ratio is set to $\alpha=0.1$ for
all models, heads and inputs, corresponding to
$k_{\mathrm{fix}}=\lceil0.1N\rceil$. The first attention
layer remains dense at every denoising step. Wan2.1 and
Wan2.2 generate 81 frames, for all remaining layers, dense
attention is retained during the first $20\%$ of denoising
steps and SCOPE is applied thereafter. HunyuanVideo
generates 129 frames and uses the same strategy with a
$10\%$ dense prefix.

\subsection{Main Results}

\begin{wrapfigure}{r}{0.44\textwidth}
    \vspace{-\baselineskip}
    \centering
    \includegraphics[width=\linewidth]{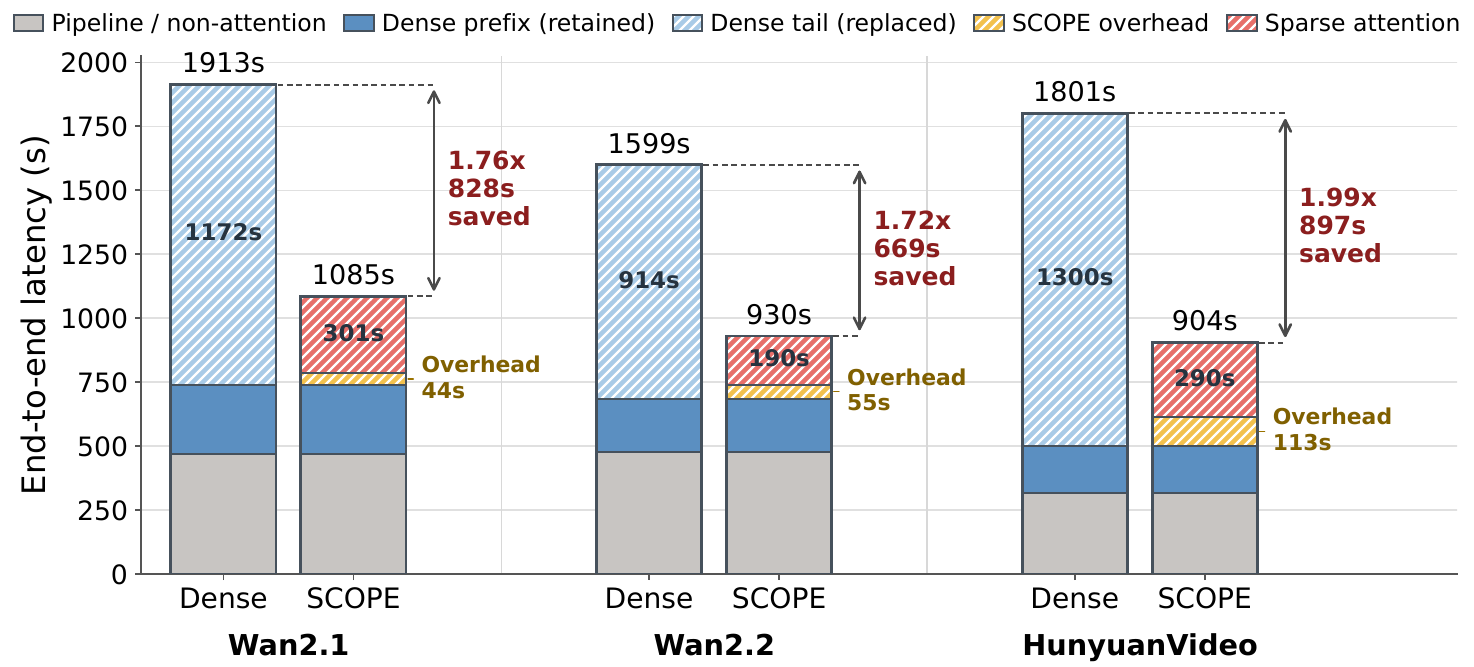}
    \captionsetup{font=scriptsize,skip=2pt}
    \caption{Runtime composition across three text-to-video models.
SCOPE retains the dense prefix and replaces the
dense-attention tail with sparse attention. The additional
overhead remains small,
resulting in $1.72\times$--$1.99\times$ end-to-end
speedups.
}
    \label{fig:latency-breakdown}
\end{wrapfigure}

\paragraph{Quality and efficiency.}
Tables~\ref{tab:main-results} and~\ref{tab:i2v-results}
compare SCOPE with training-free sparse attention baselines
across six 720p T2V and I2V settings. SCOPE is the only
method that records the lowest measured latency while ranking
first on all three dense-reference fidelity metrics (PSNR,
SSIM and LPIPS) in every setting. It achieves
$1.67\times$--$1.99\times$ end-to-end speedups, while its
VBench scores remain close to the dense-attention reference
and comparable to those of the competing methods.
Figure~\ref{fig:quality-compare} further shows that SCOPE
preserves the appearance and structure of the dense-attention
output more faithfully than the competing sparse attention
methods.

\paragraph{Runtime breakdown.}
As shown in Figure~\ref{fig:latency-breakdown}, the additional
cost of subspace clustering, proxy scoring and online key
selection is consistently outweighed by the reduction in
attention time. SCOPE therefore converts sparse attention
into substantial end-to-end acceleration across all three
T2V models.

\subsection{Ablation Studies}

\begin{wrapfigure}{r}{0.44\textwidth}
    \vspace{-\baselineskip}
    \centering
    \includegraphics[width=\linewidth]{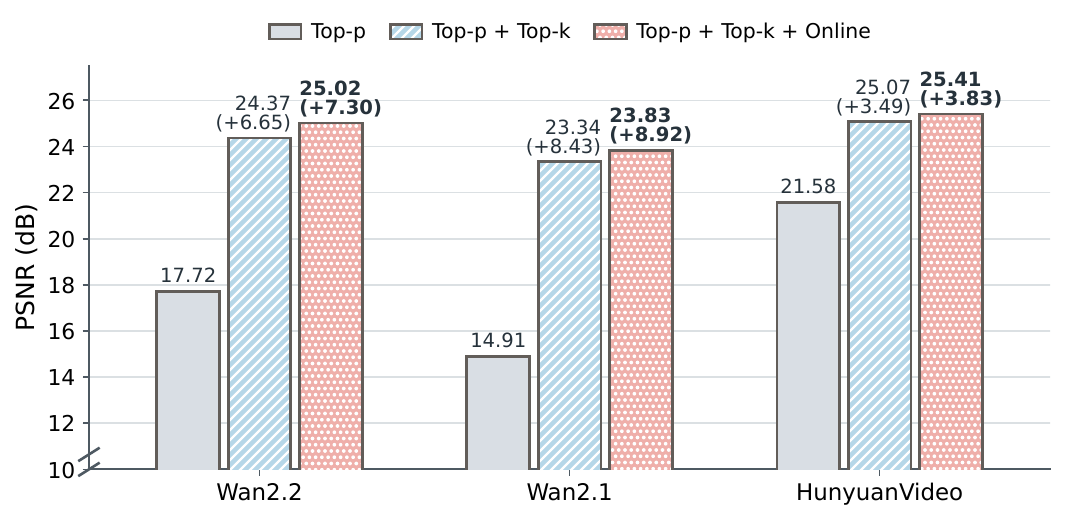}
    \captionsetup{font=scriptsize,skip=2pt}
    \caption{
Ablation of online per-head Top-$k$ estimation. We set $p$ seperately for each variant to match realized attention density($\alpha=0.1$ when Top-$k$ is used). The fixed floor mitigates Top-$p$ under-selection, while online estimation further improves PSNR on all three models.
}
    \label{fig:online-ablation}

    \vspace{8pt}
    {\small
    \setlength{\tabcolsep}{4.5pt}
    \renewcommand{\arraystretch}{1.08}
    \begin{tabular}{lcc}
        \toprule
        Partition
        & PSNR $\uparrow$
        & Latency $\downarrow$\\
        \midrule
        1 (Full-dim)       & 24.17            & 878s  \\
        2                   & 25.53            & 897s  \\
        3 (Random)          & 25.61            & 931s  \\
        \textbf{3 (3D RoPE)}
                            & \underline{25.90} & 927s  \\
        4                   & 25.72            & 948s  \\
        8                   & \textbf{26.07}   & 1011s \\
        \bottomrule
    \end{tabular}
    }
    \captionsetup{font=scriptsize,skip=2pt}
    \captionof{table}{
Ablation of key subspace partitioning.
Each subspace uses 333 centroids. ``1 (Full-dim)'' denotes
full-dimensional clustering without decomposition,
``3 (Random)'' randomly partitions the channels into three
subspaces and ``3 (3D RoPE)'' follows the temporal, height,
and width channel ranges defined by 3D RoPE.
}
    \label{tab:subspace-ablation}
\end{wrapfigure}

\paragraph{Online per-head Top-$k$ estimation.}
Figure~\ref{fig:online-ablation} compares Top-$p$, hybrid
Top-$p$/fixed-Top-$k$ and the complete online strategy
under matched realized attention density. The fixed floor strongly improves over pure Top-$p$, confirming under-selection from approximate proxy distributions. With realized density matched, the further PSNR gains from online estimation indicate better allocation across query clusters rather than a larger budget.

\paragraph{Key subspace partitioning.}
Using 333 centroids per subspace,
Table~\ref{tab:subspace-ablation} compares different
key-space partitions. Full-dimensional clustering reaches
only 24.17\,dB, while all decomposed variants improve
PSNR. The proposed 3D-RoPE-aligned partition achieves
25.90\,dB, outperforming the random three-way split by
0.29\,dB at slightly lower latency and the four-way split by
0.18\,dB while saving 21\,s. Increasing to eight subspaces
yields only another 0.17\,dB but adds 84\,s. The
temporal--height--width partition therefore offers the most
favorable fidelity--efficiency trade-off.

\par\vspace{3.5\baselineskip}
\WFclear

%% file: sections/conclusion.tex
\section{Conclusion}

We presented SCOPE, a training-free sparse attention
framework for efficient video DiT inference. SCOPE
partitions post-RoPE keys according to the temporal, height and width channel ranges of 3D RoPE and clusters the three
subspaces independently. Compositional centroid lookups
then recover proxy logits for individual keys without scoring
every key against every query centroid. SCOPE also
estimates a Top-$k$ value online for each head by averaging
the initial retained counts with weights given by query
cluster size and extends only clusters whose counts fall
below this estimate. These proxy quantities are used solely
for key selection, while attention is computed with the
original tokens. Across six 720p T2V and
I2V settings on Wan2.1, Wan2.2 and HunyuanVideo,
SCOPE achieves higher fidelity to dense attention and lower
measured latency than competing training-free methods,
delivering $1.67\times$--$1.99\times$ end-to-end speedups
while maintaining generation quality. Together, these
results validate subspace clustering and online Top-$k$
estimation as complementary mechanisms for accurate and
efficient sparse video attention.

%% file: supplementary/sections/scope_appendix_sections_A_B.tex

\section{Algorithmic Details}
\label{sec:supp_algorithm}

We provide pseudocode for SCOPE. All quantities in this section refer to one attention head at a fixed layer and denoising step. Let $Q,K\in\mathbb{R}^{N\times d}$ denote the post-3D-RoPE query and key matrices, and let $V\in\mathbb{R}^{N\times d}$ denote the value matrix. We write $\mathcal{M}=\{T,H,W\}$ for the temporal, height and width RoPE subspaces. Their channel-index sets $\{\mathcal{I}_m\}_{m\in\mathcal{M}}$ are disjoint, cover all $d$ channels and have dimensions $d_m=|\mathcal{I}_m|$, so that $\sum_{m\in\mathcal{M}}d_m=d$.

For $X\in\mathbb{R}^{n\times d_x}$, $\operatorname{KMeans}(X,C)$ returns $C$ centroids and an assignment vector $a\in\{1,\ldots,C\}^{n}$, where $a_i$ is the index of the centroid assigned to row $X[i,:]$. Full-dimensional query clustering therefore produces $\bar Q\in\mathbb{R}^{C_q\times d}$ and $g\in\{1,\ldots,C_q\}^{N}$, where row $c$ of $\bar Q$ is $\bar q_c$. We define $\mathcal{Q}_c=\{i:g_i=c\}$ and $n_c=|\mathcal{Q}_c|$. Algorithm~\ref{alg:scope_full} summarizes the complete per-head SCOPE pipeline.

\begin{algorithm}[H]
\caption{SCOPE for one attention head}
\label{alg:scope_full}
\begin{algorithmic}[1]
\Require Post-RoPE queries $Q\in\mathbb{R}^{N\times d}$; post-RoPE keys $K\in\mathbb{R}^{N\times d}$; values $V\in\mathbb{R}^{N\times d}$; channel ranges $\{\mathcal{I}_m\}_{m\in\mathcal{M}}$; query-cluster count $C_q$; key-codebook sizes $\{C_m\}_{m\in\mathcal{M}}$; Top-$p$ threshold $\rho$; fixed Top-$k$ ratio $\alpha$
\Ensure Output $O\in\mathbb{R}^{N\times d}$ and selected key sets $\{\mathcal{S}_c\}_{c=1}^{C_q}$
\State $(\bar Q,g)\gets\operatorname{KMeans}(Q,C_q)$
\For{$c=1,\ldots,C_q$}
    \State $\mathcal{Q}_c\gets\{i:g_i=c\}$; $n_c\gets|\mathcal{Q}_c|$
\EndFor
\For{$m\in\mathcal{M}$}
    \State $\bar Q^m\gets\bar Q[:,\mathcal{I}_m]$; $K^m\gets K[:,\mathcal{I}_m]$
\EndFor
\State $\widetilde S\gets\operatorname{KeySubspaceScoring}\!\left(\{\bar Q^m,K^m,C_m\}_{m\in\mathcal{M}}\right)$
\State $\{\mathcal{S}_c\}_{c=1}^{C_q}\gets\operatorname{OnlinePerHeadTopK}(\widetilde S,\{n_c\}_{c=1}^{C_q},\rho,\alpha)$
\For{$c=1,\ldots,C_q$}
    \State $O[\mathcal{Q}_c,:]\gets\operatorname{softmax}\!\left(\dfrac{Q[\mathcal{Q}_c,:]K[\mathcal{S}_c,:]^{\top}}{\sqrt d}\right)V[\mathcal{S}_c,:]$
\EndFor
\State \Return $O$ and $\{\mathcal{S}_c\}_{c=1}^{C_q}$
\end{algorithmic}
\end{algorithm}

\subsection{3D-RoPE-Aligned Key Subspace Scoring}

Algorithm~\ref{alg:scope_score} details the key-side operations in Steps 1--2 of the method overview in the main paper. It independently clusters the temporal, height and width key subspaces and then constructs the corresponding query-centroid--key-centroid score tables. For subspace $m$, the centroid matrix is $U^m\in\mathbb{R}^{C_m\times d_m}$, and $z_j^m\in\{1,\ldots,C_m\}$ is the centroid assignment of key $j$. The three assignments form the composite code $\mathbf z_j=(z_j^T,z_j^H,z_j^W)$ defined in the main paper.

The input to Algorithm~\ref{alg:scope_score} consists of the sliced matrices $\bar Q^m$ and $K^m$. K-means is applied to $K^m$ to obtain $U^m$ and $z^m$, whereas proxy-score dot products are computed between $\bar Q^m$ and the key-subspace centroids $U^m$:
$P_m=\bar Q^m(U^m)^{\top}$. Thus, the proxy-scoring stage does not directly multiply $\bar Q$ by the original key matrix $K$.

For $z^m=(z_1^m,\ldots,z_N^m)$, the notation
$P_m[:,z^m]\in\mathbb{R}^{C_q\times N}$ denotes the matrix
whose $j$-th column is $P_m[:,z_j^m]$. Adding
$P_m[:,z^m]$ therefore applies the indexed proxy-score
lookup to all keys simultaneously.
\subsection{Online Per-Head Top-$k$ Selection}

Algorithm~\ref{alg:scope_select} implements Step 3 of the method overview in the main paper. It first computes the base retained count $b_c$ from hybrid Top-$p$/fixed-Top-$k$ selection. Because both selections are prefixes of the same descending ranking $\pi_c$, their union contains exactly $b_c=\max\{t_c^{(p)},k_{\mathrm{fix}}\}$ keys. The online stage then computes a query-size-weighted head-level minimum and extends only the clusters below that minimum, without constructing another token-level ranking.

The weighting by $n_c$ makes the estimator a per-query average rather than an unweighted per-cluster average. In particular,
\begin{equation}
 k_{\mathrm{head}}
 =\left\lceil\frac{1}{N}\sum_{i=1}^{N}b_{g_i}\right\rceil
 =\left\lceil\frac{\sum_{c=1}^{C_q}n_cb_c}{\sum_{c=1}^{C_q}n_c}\right\rceil.
\end{equation}
The cluster-based query permutation shown in the main-paper
overview only groups queries with the same assignment $g_i$.
Algorithm~\ref{alg:scope_full} writes the same computation
directly using the index sets $\mathcal{Q}_c$, so no separate
permutation variable is required.

\begin{center}
\begin{minipage}[t]{0.48\textwidth}
\begin{algorithm}[H]
\caption{3D-RoPE-aligned key subspace clustering and scoring}
\label{alg:scope_score}
\begin{algorithmic}[1]
\Require Query-centroid slices $\{\bar Q^m\in\mathbb{R}^{C_q\times d_m}\}_{m\in\mathcal{M}}$; post-RoPE key subspaces $\{K^m\in\mathbb{R}^{N\times d_m}\}_{m\in\mathcal{M}}$; codebook sizes $\{C_m\}_{m\in\mathcal{M}}$
\Ensure Proxy-logit matrix $\widetilde S\in\mathbb{R}^{C_q\times N}$
\State $\widetilde S\gets 0\in\mathbb{R}^{C_q\times N}$
\For{$m\in\mathcal{M}$}
    \State $(U^m,z^m)\gets\operatorname{KMeans}(K^m,C_m)$
    \State $P_m\gets\bar Q^m(U^m)^{\top}$ \Comment{$P_m\in\mathbb{R}^{C_q\times C_m}$}
    \State $\widetilde S\gets\widetilde S+P_m[:,z^m]$
\EndFor
\State $\widetilde S\gets\widetilde S/\sqrt d$
\State \Return $\widetilde S$
\end{algorithmic}
\end{algorithm}

For query cluster $c$ and key $j$, Algorithm~\ref{alg:scope_score} computes
\begin{equation}
\widetilde s_{c,j}
=\frac{1}{\sqrt d}\sum_{m\in\mathcal{M}}P_m[c,z_j^m],
\end{equation}
which is the proxy-logit definition in the main method. The score tables, subspace assignments and proxy logits are used only to construct the selected key sets, sparse attention subsequently uses the original $Q$, $K$ and $V$.
\end{minipage}
\hfill
\begin{minipage}[t]{0.48\textwidth}
\begin{algorithm}[H]
\caption{Online per-head Top-$k$ selection}
\label{alg:scope_select}
\begin{algorithmic}[1]
\Require Proxy logits $\widetilde S\in\mathbb{R}^{C_q\times N}$; query-cluster sizes $\{n_c\}_{c=1}^{C_q}$; Top-$p$ threshold $\rho\in(0,1]$; fixed Top-$k$ ratio $\alpha\in(0,1]$
\Ensure Selected key sets $\{\mathcal{S}_c\}_{c=1}^{C_q}$
\State $N\gets\operatorname{ncols}(\widetilde S)$; $k_{\mathrm{fix}}\gets\lceil\alpha N\rceil$
\For{$c=1,\ldots,C_q$}
    \State $\pi_c\gets\operatorname*{argsort}_{\downarrow}(\widetilde S[c,:])$
    \State $\widetilde a_c\gets\operatorname{softmax}(\widetilde S[c,:])$
    \State $t_c^{(p)}\gets\min\left\{t:\sum_{\ell=1}^{t}\widetilde a_{c,\pi_c(\ell)}\ge\rho\right\}$
    \State $b_c\gets\max\{t_c^{(p)},k_{\mathrm{fix}}\}$
\EndFor
\State $k_{\mathrm{head}}\gets\left\lceil\dfrac{\sum_{c=1}^{C_q}n_cb_c}{\sum_{c=1}^{C_q}n_c}\right\rceil$
\For{$c=1,\ldots,C_q$}
    \State $r_c\gets\max\{b_c,k_{\mathrm{head}}\}$
    \State $\mathcal{S}_c\gets\{\pi_c(1),\ldots,\pi_c(r_c)\}$
\EndFor
\State \Return $\{\mathcal{S}_c\}_{c=1}^{C_q}$
\end{algorithmic}
\end{algorithm}
\end{minipage}
\end{center}
\FloatBarrier

\section{Complexity Analysis}
\label{sec:supp_complexity}

We analyze the compositional key-scoring module for one
attention head, treating the query centroids, key-subspace
codebooks and key assignments as its inputs. This isolates
SCOPE's central computation--representation trade-off: the
high-dimensional scoring cost is determined by an additive
sum over subspaces, whereas the available compositional code
space grows multiplicatively. Let $\bar Q\in\mathbb{R}^{C_q\times d}$ contain the $C_q$ query centroids. The key channels are partitioned into $M$ disjoint subspaces with dimensions $d_1,\ldots,d_M$, where $\sum_{m=1}^{M}d_m=d$. Subspace $m$ contains $C_m$ centroids $U^m\in\mathbb{R}^{C_m\times d_m}$ and an assignment vector $z^m$ for the $N$ keys. SCOPE uses $M=3$ for the temporal, height and width ranges of 3D RoPE.

\subsection{Compositional Scoring Cost}

Directly scoring every key against every query centroid evaluates
\begin{equation}
S^{\star}=\frac{\bar QK^{\top}}{\sqrt d},
\qquad
S^{\star}\in\mathbb{R}^{C_q\times N},
\end{equation}
and requires
\begin{equation}
T_{\mathrm{direct}}^{\mathrm{MAC}}=C_qNd
\label{eq:supp_direct_scoring}
\end{equation}
high-dimensional multiply--accumulate operations (MACs).

SCOPE instead constructs one compact score table per key subspace,
\begin{equation}
P_m=\bar Q^m(U^m)^{\top},
\qquad
P_m\in\mathbb{R}^{C_q\times C_m}.
\end{equation}
The total table-construction cost is
\begin{equation}
T_{\mathrm{table}}^{\mathrm{MAC}}
=C_q\sum_{m=1}^{M}C_md_m.
\label{eq:supp_table_scoring}
\end{equation}
The proxy logits are then expanded through indexed accumulation,
\begin{equation}
\widetilde S
=\frac{1}{\sqrt d}\sum_{m=1}^{M}P_m[:,z^m],
\label{eq:supp_proxy_reconstruction}
\end{equation}
which requires $M$ table lookups and $M-1$ scalar additions per query-cluster--key pair. Hence, the $d$-dimensional dot products are confined to the compact centroid tables, while all per-key proxy scores are recovered using scalar operations.

When all subspaces use the same codebook size $C_m=C$,
\begin{equation}
T_{\mathrm{table}}^{\mathrm{MAC}}
=C_qC\sum_{m=1}^{M}d_m
=C_qCd.
\label{eq:supp_equal_table_scoring}
\end{equation}
Relative to direct query-centroid--key scoring, the high-dimensional scoring cost is reduced by
\begin{equation}
\frac{T_{\mathrm{direct}}^{\mathrm{MAC}}}
     {T_{\mathrm{table}}^{\mathrm{MAC}}}
=\frac{N}{C}.
\label{eq:supp_scoring_reduction}
\end{equation}

\begin{figure}[t]
    \centering
    \includegraphics[width=\textwidth]{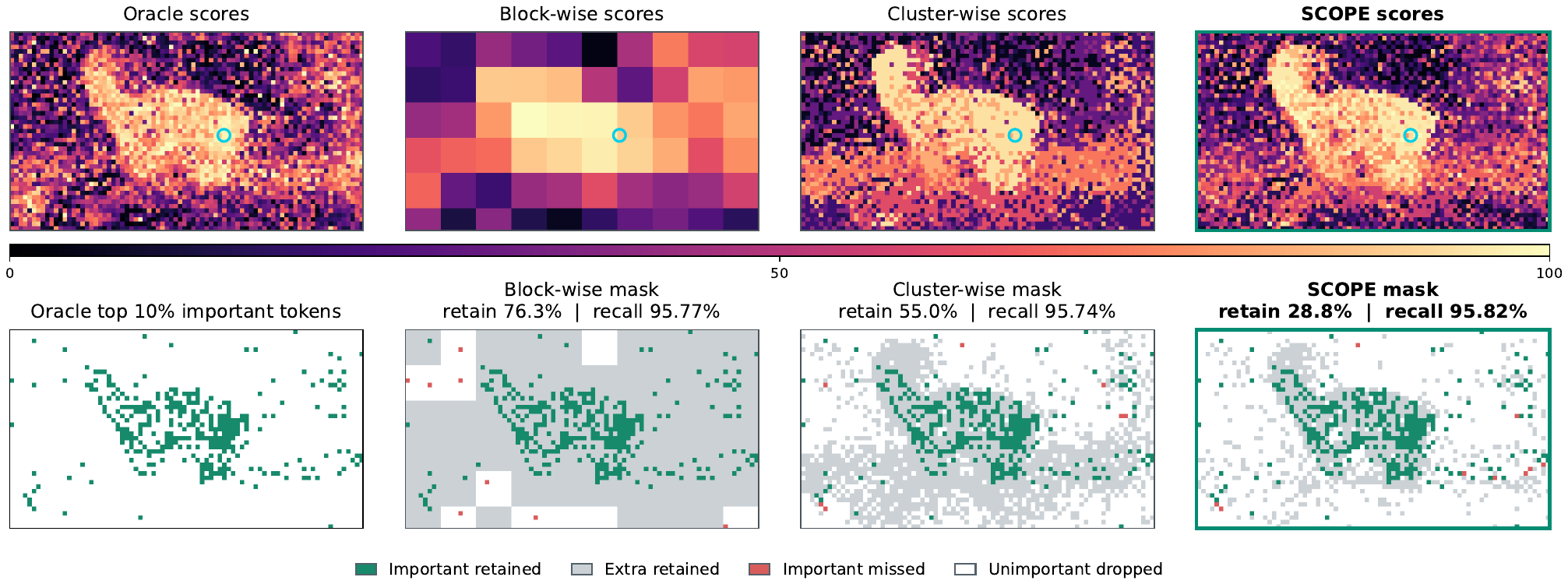}
    \caption{
Fine-grained key selection at matched oracle recall.
For a representative query token marked by the cyan circle,
the top row compares exact and proxy query--key score maps,
while the bottom row compares the oracle top-$10\%$ key set
with the selected masks. At approximately $95.8\%$ oracle
recall, SCOPE retains only $28.8\%$ of all keys, compared
with $55.0\%$ for full-dimensional cluster-wise scoring and
$76.3\%$ for block-wise scoring.
}
    \label{fig:supp_matched_recall}
\end{figure}

\subsection{Additive Scoring Cost and Multiplicative Code Space}

Each key is represented by the composite assignment
\begin{equation}
\mathbf z_j=(z_j^1,\ldots,z_j^M).
\end{equation}
The number of centroid scores required for each query cluster grows additively with the subspace codebook sizes,
\begin{equation}
C_{\mathrm{score}}=\sum_{m=1}^{M}C_m,
\end{equation}
whereas the number of available composite key codes grows multiplicatively,
\begin{equation}
C_{\mathrm{code}}=\prod_{m=1}^{M}C_m.
\label{eq:supp_product_code_space}
\end{equation}
For equal codebook sizes,
\begin{equation}
C_{\mathrm{score}}=MC,
\qquad
C_{\mathrm{code}}=C^M,
\qquad
\sum_{m=1}^{M}C_md_m=Cd.
\label{eq:supp_additive_multiplicative}
\end{equation}
Thus, a single full-dimensional codebook with $C$ centroids and the SCOPE product code require the same centroid storage, $Cd$, and the same high-dimensional score-table MAC count, $C_qCd$. However, the full-dimensional codebook provides only $C$ assignment labels, whereas the product construction provides up to $C^M$ structured assignment tuples. At matched centroid storage and high-dimensional scoring cost, SCOPE enlarges the available code space by
\begin{equation}
\frac{C^M}{C}=C^{M-1}.
\label{eq:supp_code_space_gain}
\end{equation}
By comparison, a monolithic full-dimensional codebook with
$C^M$ centroids would require $C^M d$ centroid scalars and
$C_q C^M d$ MACs to construct its score table, whereas the
product construction requires only $Cd$ centroid scalars and
$C_qCd$ score-table MACs. Here, $C^M$ denotes the
cardinality of the available structured code space, at most
$\min\{N,C^M\}$ distinct codes can be instantiated by the
observed keys, and distinct codes need not yield distinct
scalar proxy logits for a given query centroid.

Finally, define the product-code representative of key $j$ as
\begin{equation}
\widehat k_j
=\operatorname{concat}\!\left(
 u^1_{z_j^1},\ldots,u^M_{z_j^M}
\right).
\end{equation}
Because the subspaces occupy disjoint channel ranges,
\begin{equation}
\widetilde s_{c,j}
=\frac{1}{\sqrt d}\sum_{m=1}^{M}
\left\langle \bar q_c^m,u^m_{z_j^m}\right\rangle
=\frac{\langle\bar q_c,\widehat k_j\rangle}{\sqrt d}.
\label{eq:supp_lookup_exactness}
\end{equation}
Thus, the lookup-and-sum operation exactly evaluates the
inner product with the product-code representative, it
introduces no approximation beyond replacing each key with
that representative.

\section{Additional Quantitative Results}

\subsection{Fine-Grained Key Selection at Matched Oracle Recall}
\label{sec:supp_matched_recall}

Figure~\ref{fig:supp_matched_recall} compares the
key-selection granularity of different proxy scoring schemes
for a representative query token. The exact query--key scores
define the oracle top-$10\%$ key set. For each proxy, keys are
ranked by their estimated scores and retained until
approximately the same oracle recall is reached. Retention
denotes the fraction of all key tokens selected, whereas
oracle recall denotes the fraction of oracle-important keys
recovered.

Block-wise scoring assigns one proxy score to all keys in the
same block, so retaining an important key may also retain many
unimportant keys from that block. Full-dimensional
cluster-wise scoring provides a finer grouping, but all keys
assigned to the same centroid still share one score and tend
to be retained together. SCOPE instead combines temporal,
height and width subspace assignments to construct more
fine-grained proxy scores. Keys that share a representative in
one subspace can therefore still be distinguished by their
assignments in the other subspaces.

All three methods recover approximately $95.8\%$ of the
oracle-important keys. SCOPE achieves this recall while
retaining only $28.8\%$ of all key tokens, compared with
$55.0\%$ for full-dimensional cluster-wise scoring and
$76.3\%$ for block-wise scoring. The substantially lower
retention ratio shows that SCOPE reduces the over-selection
of unimportant keys at comparable oracle recall.

\FloatBarrier

\section{Additional Qualitative Results}
\label{sec:supp_qualitative}

We provide additional dense-reference visual comparisons
across all six 720p model--task configurations. All
subsequent figures use the same layout. Each pair of rows
corresponds to one generation example, with Dense Attention
in the upper row and SCOPE in the lower row, the three
columns show generated frames sampled at matched temporal
positions. The paired outputs use identical prompts,
conditioning images when applicable, random seeds and
sampling configurations.

\begin{figure}[!htbp]
    \centering
    \begin{minipage}{\textwidth}
        \centering
        \captionsetup{skip=0pt}
        \includegraphics[
            width=0.82\textwidth
        ]{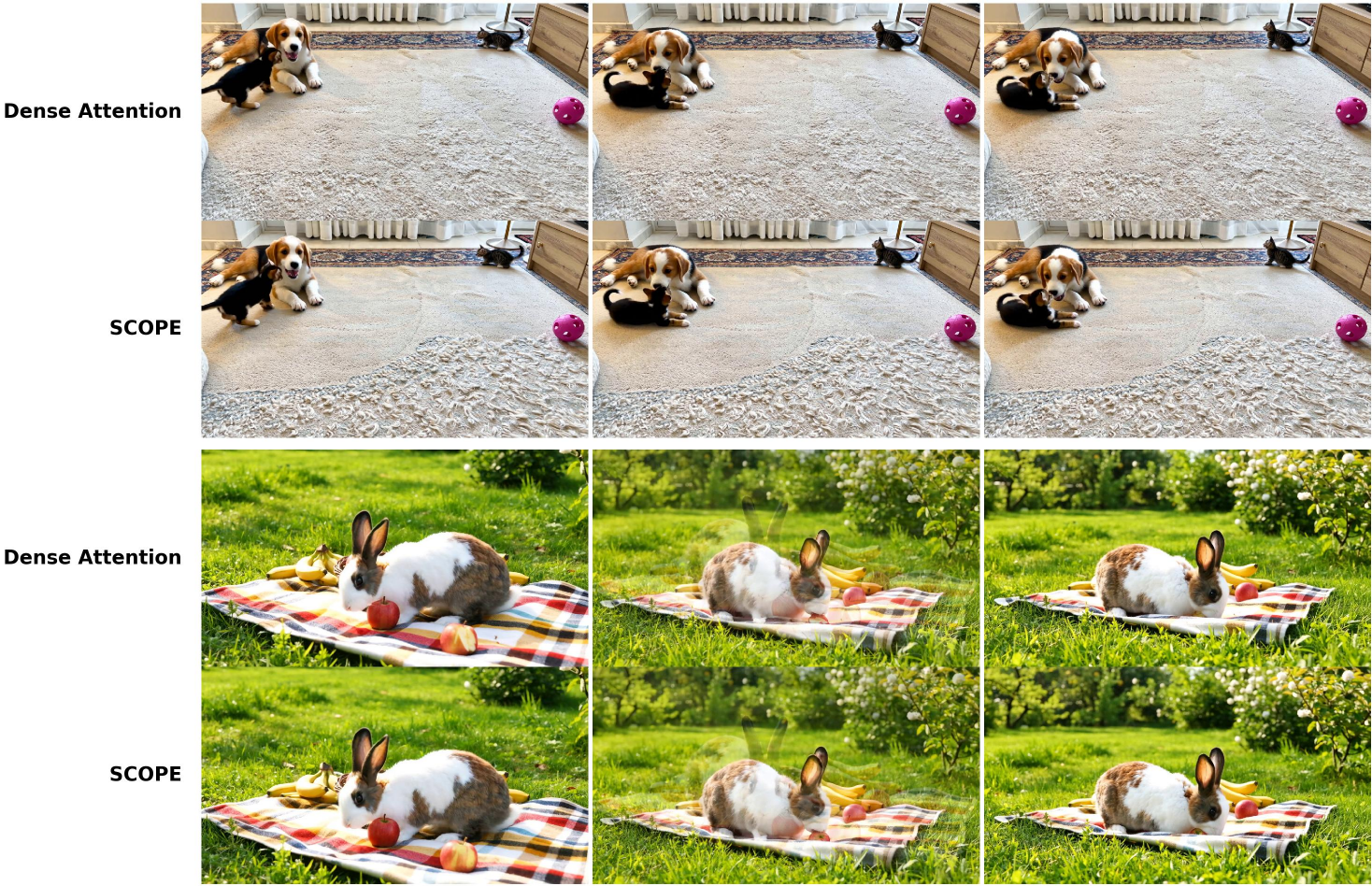}
        \captionof{figure}{
        Additional 720p text-to-video comparisons on
        Wan2.2-A14B.
        }
        \label{fig:supp_wan22_t2v}

        \includegraphics[
            width=0.82\textwidth
        ]{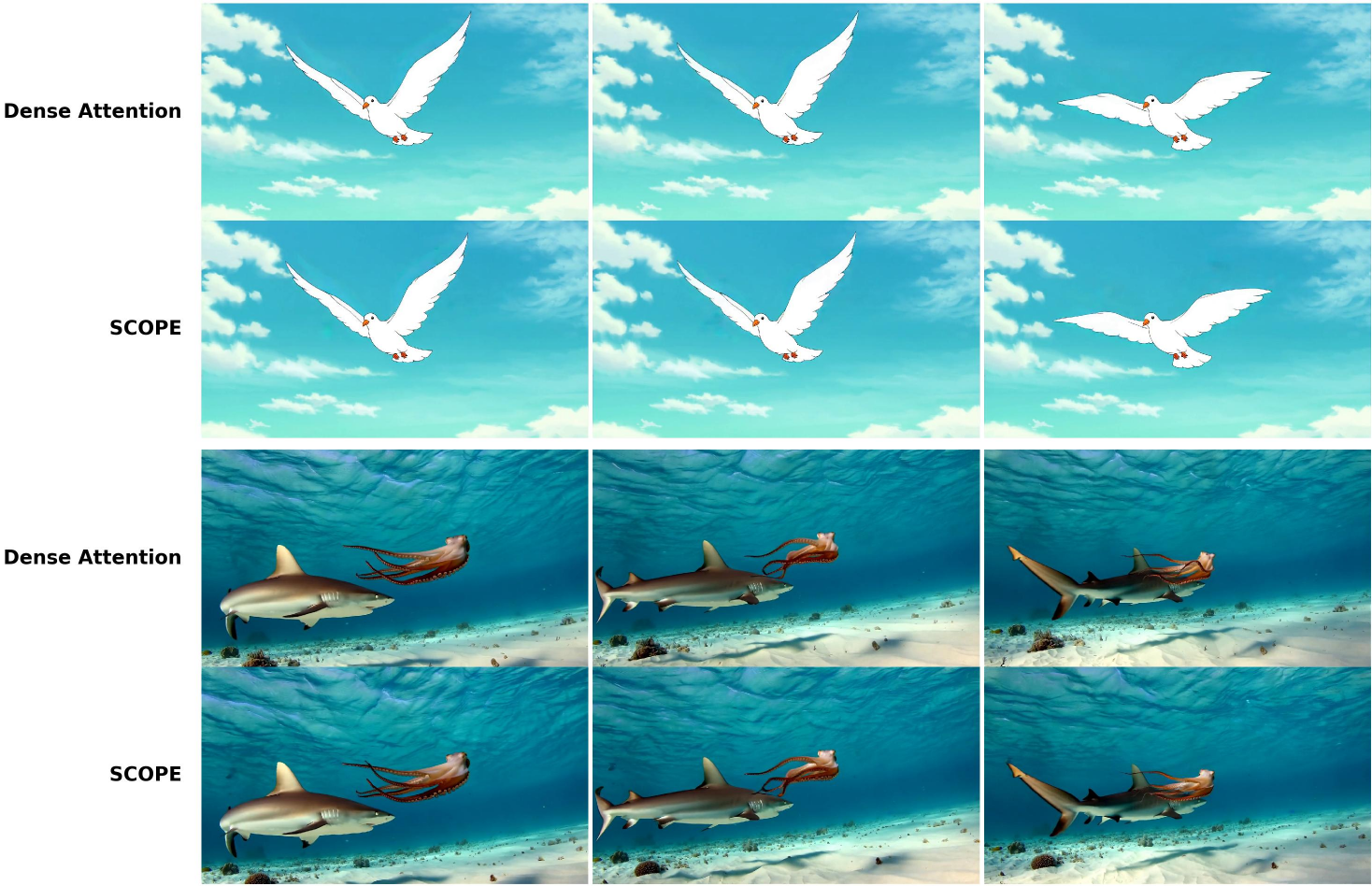}
        \captionof{figure}{
        Additional 720p text-to-video comparisons on
        Wan2.1-14B.
        }
        \label{fig:supp_wan21_t2v}
    \end{minipage}
\end{figure}

\begin{figure}[!p]
    \centering
    \includegraphics[
        width=0.82\textwidth
    ]{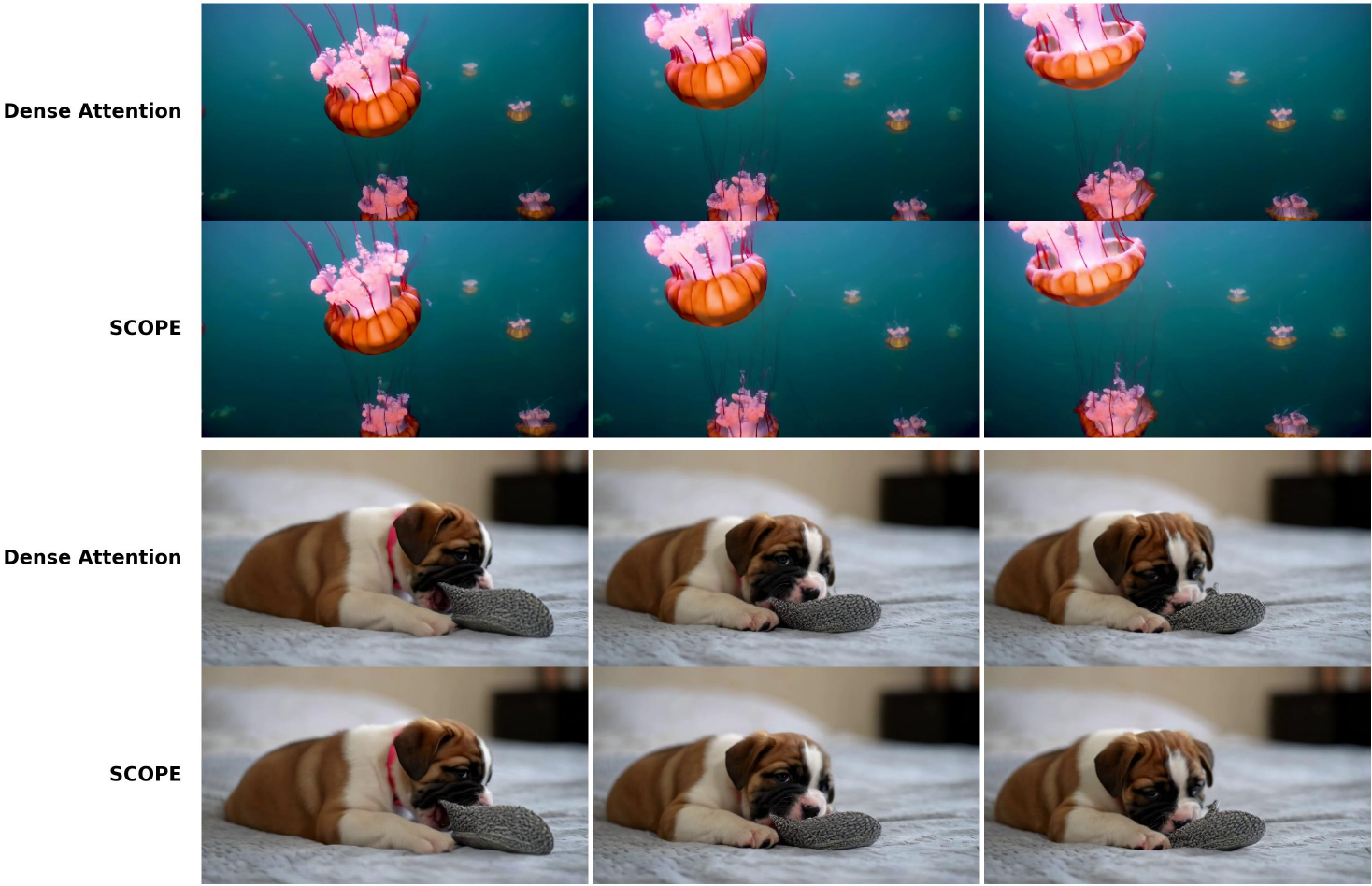}
    \caption{
    Additional 720p text-to-video comparisons on
    HunyuanVideo-13B.
    }
    \label{fig:supp_hunyuan_t2v}

    \includegraphics[
        width=0.82\textwidth
    ]{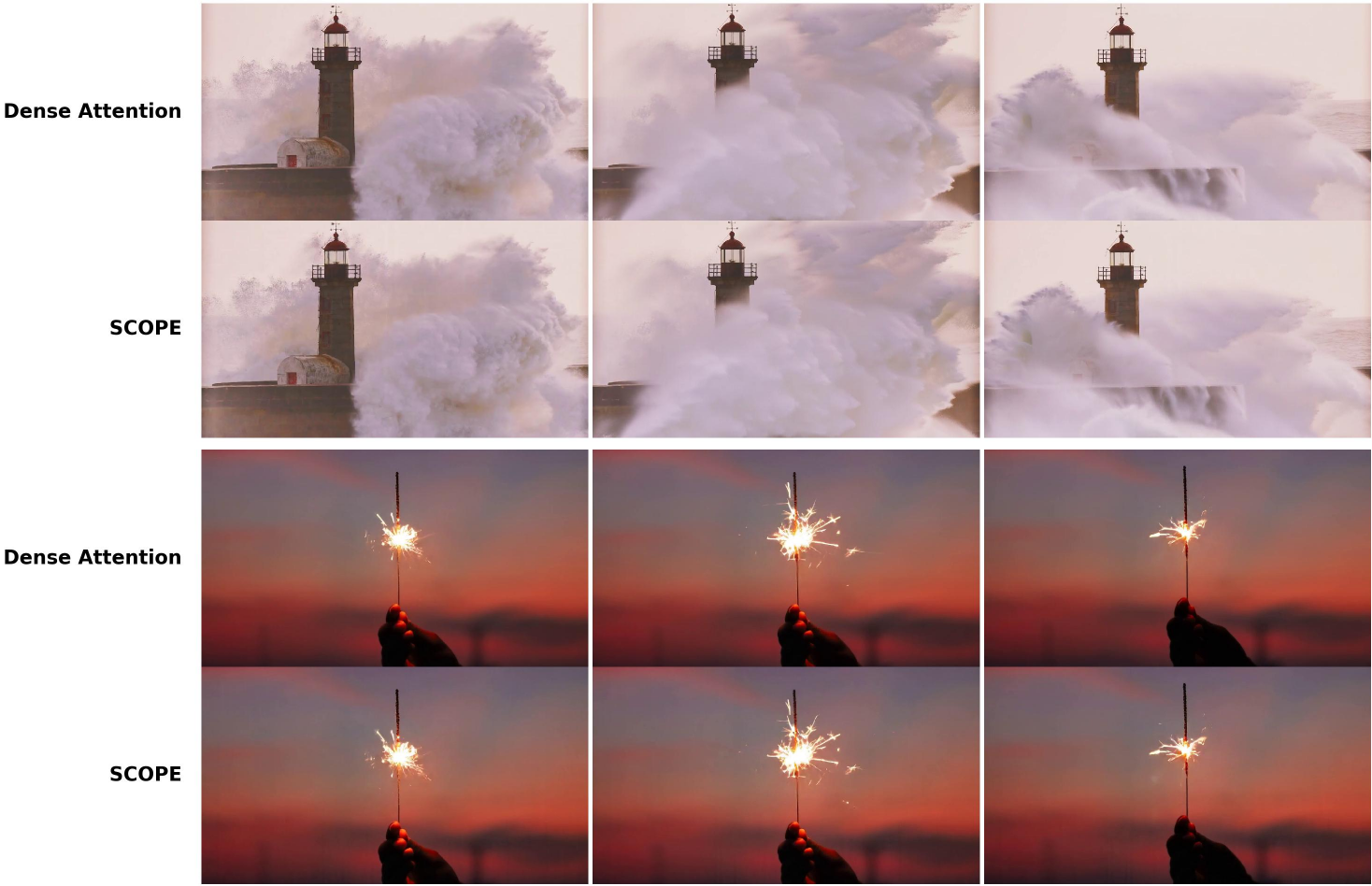}
    \caption{
    Additional 720p image-to-video comparisons on
    Wan2.2-A14B.
    }
    \label{fig:supp_wan22_i2v}
\end{figure}

\begin{figure}[!p]
    \centering
    \includegraphics[
        width=0.82\textwidth
    ]{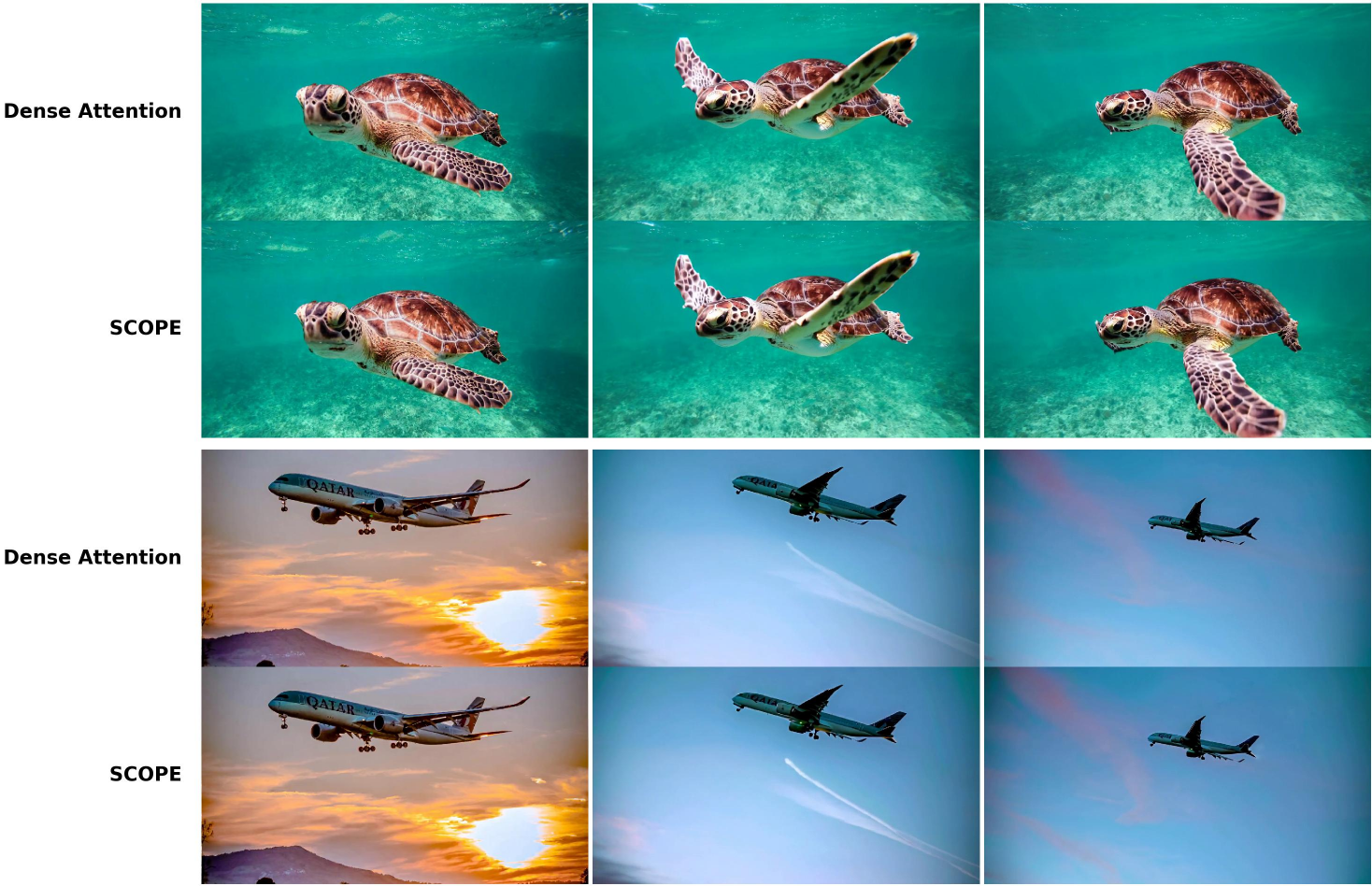}
    \caption{
    Additional 720p image-to-video comparisons on
    Wan2.1-14B.
    }
    \label{fig:supp_wan21_i2v}

    \includegraphics[
        width=0.82\textwidth
    ]{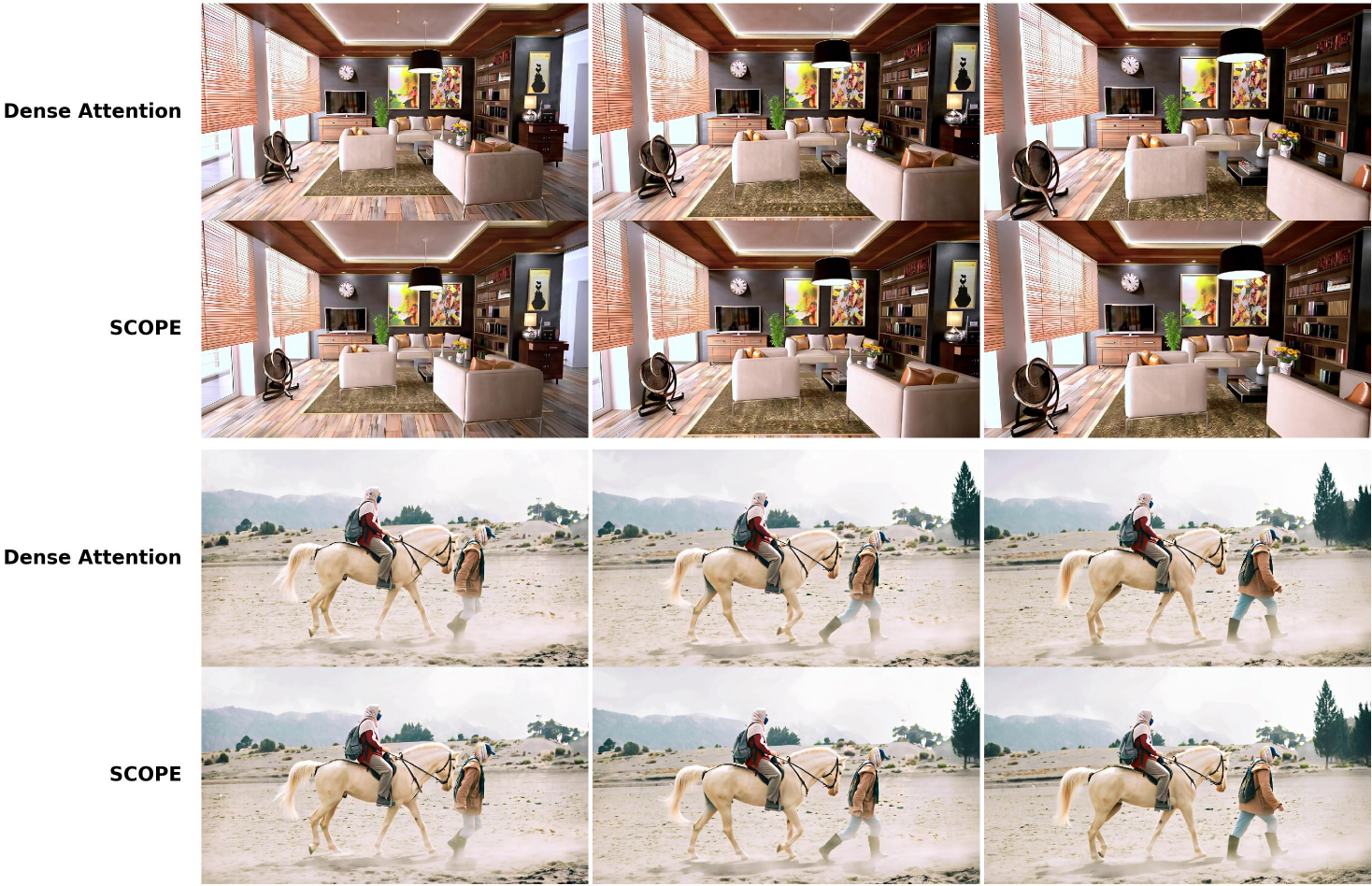}
    \caption{
    Additional 720p image-to-video comparisons on
    HunyuanVideo-13B.
    }
    \label{fig:supp_hunyuan_i2v}
\end{figure}